\documentclass[10pt,twocolumn,letterpaper]{article}

\usepackage{cvpr}              
\usepackage{multirow} 
\definecolor{cvprblue}{rgb}{0.21,0.49,0.74}
\usepackage[pagebackref,breaklinks,colorlinks,allcolors=cvprblue]{hyperref}

\def\paperID{} 
\def\confName{CVPR}
\def\confYear{2026}

\title{ClimaOoD:  Improving Anomaly Segmentation via Physically Realistic Synthetic Data}

\author{
Yuxing Liu$^{1}$ \quad
Zheng Li$^{1}$ \quad
Huanhuan Liang$^{1}$ \quad
Ji Zhang$^{2\dagger}$ \quad
Zeyu Sun$^{3}$ \quad
Yong Liu$^{1\dagger}$\\[0.5em]
$^{1}$Beijing University of Chemical Technology \\
$^{2}$Southwest Minzu University \\
$^{3}$Institute of Software, Chinese Academy of Sciences\\[0.5em]
{\tt\small lyxlyx\_47@outlook.com \quad lizheng@mail.buct.edu.cn \quad Lianghh717@gmail.com}\\
{\tt\small jizhang901@gmail.com \quad zeyu.zys@gmail.com \quad lyong@mail.buct.edu.cn}\\[0.25em]
{\tt\small $\dagger$Corresponding authors}
}

\begin{document}

\twocolumn[{%
\renewcommand \twocolumn[1][]{#1}%
\maketitle

\begin{center}
    \centering
    \vspace{-3mm}
    \includegraphics[width=\textwidth]{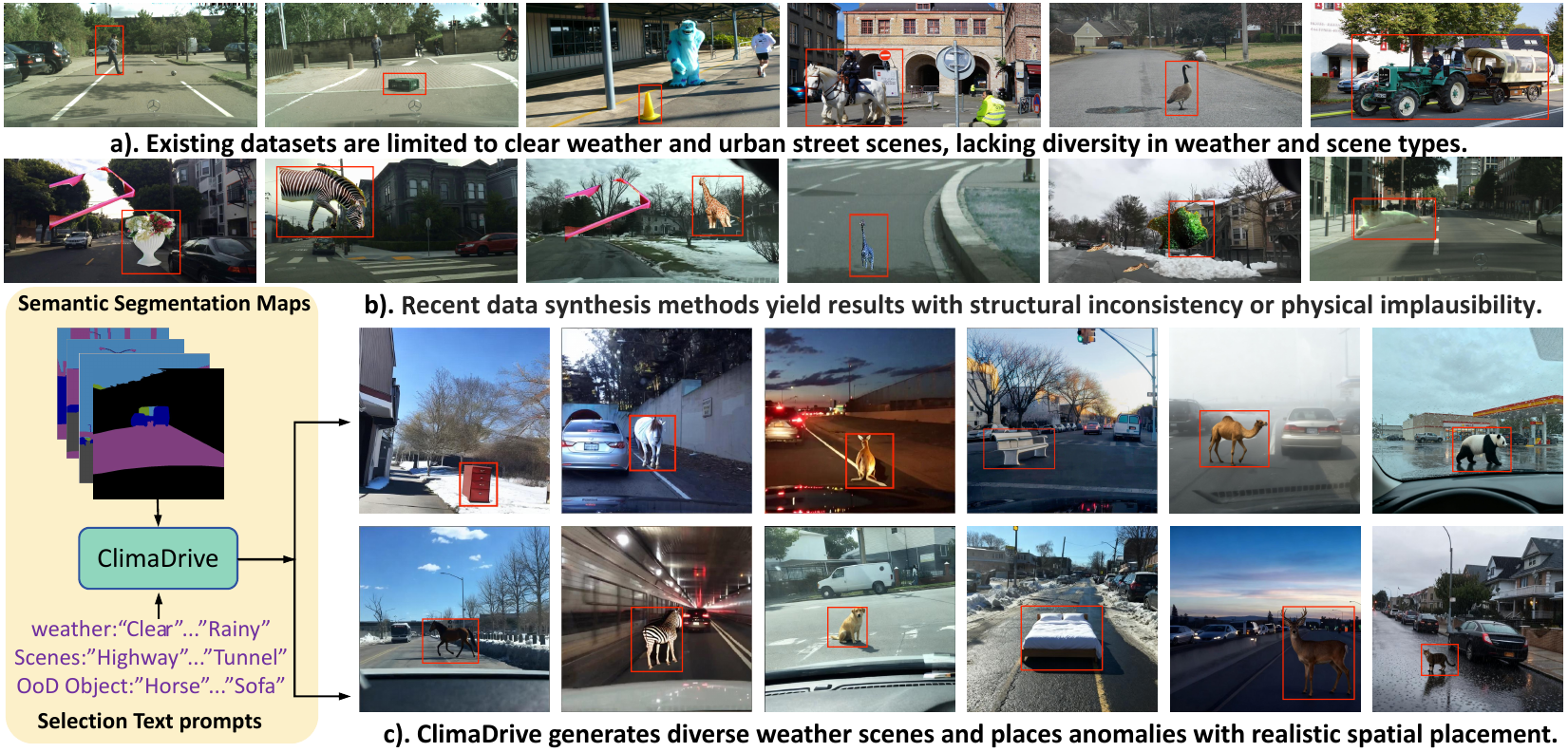}
    \captionof{figure}{Visual comparison of dataset limitations and ClimaOoD. (a) Existing datasets suffer from limited weather and scene diversity. (b) Recent data synthesis leads to contextual inconsistencies and physically unrealistic object placement. (c) ClimaDrive generates diverse weather scenes and places anomalies with realistic spatial arrangement, using semantic maps and text prompts for better OoD object placement.}
    \vspace{-5mm}
    \label{fig:climadrive_comparison}
\end{center}%
}]
\begin{abstract}
Anomaly segmentation seeks to detect and localize unknown or out-of-distribution (OoD) objects that fall outside predefined semantic classes—a capability essential for safe autonomous driving. However, the scarcity and limited diversity of anomaly data severely constrain model generalization in open-world environments. Existing approaches mitigate this issue through synthetic data generation, either by copy-pasting external objects into driving scenes or by leveraging text-to-image diffusion models to inpaint anomalous regions. While these methods improve anomaly diversity, they often lack contextual coherence and physical realism, resulting in domain gaps between synthetic and real data. 
In this paper, we present ClimaDrive, a semantics-guided image-to-image framework for synthesizing semantically coherent, weather-diverse, and physically plausible OoD driving data. ClimaDrive unifies structure-guided multi-weather generation with prompt-driven anomaly inpainting, enabling the creation of visually realistic training data. Based on this framework, we construct ClimaOoD, a large-scale benchmark spanning six representative driving scenarios under both clear and adverse weather conditions.
Extensive experiments on four state-of-the-art methods show that training with ClimaOoD leads to robust improvements in anomaly segmentation. Across all methods, AUROC, AP, and FPR95 show notable gains, with FPR95 dropping from 3.97 to 3.52 for RbA on Fishyscapes LAF. These results demonstrate that ClimaOoD enhances model robustness, offering valuable training data for better generalization in open-world anomaly detection.

\end{abstract}

\section{Introduction\label{sec:intro}}

Anomaly segmentation aims to detect out-of-distribution (OoD) objects not predefined, serving as a complementary capability to semantic segmentation methods\cite{liu2023residual,nayal2023rba,lis2019detecting,rai2024mask2anomaly,blum2021fishyscapes}. This capability is crucial for safety-critical applications such as autonomous driving, where models must respond reliably to unexpected hazards like fallen cargo, construction equipment, or animals on the road\cite{djeumou2025onemodel,chen2025solve,xu2025vlmad,DiBiase2021pixelwise}. Despite notable advances in deep segmentation networks, their performance in open-world environments remains highly constrained by the availability and diversity of anomaly data\cite{chan2021segmentmeifyoucan,xia2025opensetanomalysegmentationcomplex,blum2021fishyscapes,pinggera2016lost,chan2021entropy,delić2024outlierdetectionensemblinguncertainty,lis2019detecting}.

In the real world, anomalous events are inherently rare and unpredictable, which makes collecting large-scale, high-quality anomaly data for training anomaly segmentation models extremely difficult\cite{zhang2023improving,xia2025opensetanomalysegmentationcomplex,nekrasov2025spotting,blum2021fishyscapes}. Existing benchmarks often focus on urban streets under clear weather, overlooking complex environmental conditions (e.g., tunnels, highways, rain, fog, or snow)\cite{pinggera2016lost,blum2021fishyscapes,chan2021segmentmeifyoucan,lis2019detecting} (see Figure~\ref{fig:climadrive_comparison}.a). Consequently, models trained on these datasets struggle to generalize when faced with unseen weather or novel environments\cite{rai2024mask2anomaly,zhang2023improving}.

To alleviate the scarcity of anomaly data, recent state-of-the-art methods have explored two main paradigms of synthetic data generation. The first is the copy-paste approach \cite{liu2023residual,zhao2024segment,nayal2023rba}, which crops objects from external datasets (e.g., COCO \cite{lin2014microsoft}, ADE20K \cite{zhou2019semantic}) and pastes them into driving scenes (e.g., Cityscapes \cite{cordts2016cityscapes}) to simulate OoD events. The second leverages text-to-image diffusion models \cite{loiseau2024reliability,xia2025opensetanomalysegmentationcomplex}, where masked regions in driving images are filled with anomalous objects guided by textual prompts. While both approaches enrich anomaly diversity, they remain limited in complementary ways: copy-paste methods struggles to maintain \textbf{contextual consistency}, often causing significant discrepancies in brightness and color between pasted objects and backgrounds. Meanwhile, text-to-image diffusion-based approaches lack \textbf{physical realism}, leading to implausible object placement regarding positional coherence and scale (see Figure~\ref{fig:climadrive_comparison}.b). These issues create domain gaps between synthetic training data and real-world application scenarios, ultimately compromising model robustness and generalization capability.

In fact, the image-to-image paradigm effectively mitigates contextual inconsistencies between inserted objects and surrounding scenes through its encoding–decoding process. Moreover, guided by semantic priors, it enables the generation of objects with appropriate location and scale, thereby adhering to physical constraints. 
Inspired by these observations, we propose ClimaDrive, a unified, semantics-guided image-to-image framework that redefines OoD data synthesis. Unlike previous methods that heuristically inject anomalies or rely on unconstrained inpainting, ClimaDrive integrates perspective-aware spatial constraints and semantics-consistent scene rendering directly into the generation process. This approach combines structural priors (e.g., road geometry, depth cues) with semantic distribution constraints to place anomalies accurately in context, ensuring both realistic spatial placement and physical plausibility. The multi-scene weather generator further enhances ClimaDrive by simulating diverse weather conditions (e.g., rain, snow, fog), improving environmental realism and producing semantically consistent, visually varied driving scenes—capabilities that prior methods cannot achieve.


Built upon ClimaDrive, we construct ClimaOoD, a large-scale benchmark collected to evaluate anomaly segmentation robustness in open-world conditions. ClimaOoD synthesizes over 10,000 image-mask pairs across multiple weather conditions (clear, rain, fog, snow, cloudy, nighttime) for training. After careful curation, we create a refined test set of 1,200 image-mask pairs, covering six distinct driving scenarios (e.g., city streets, highways, tunnels) and 93 anomaly types.
In comparison to existing anomaly segmentation benchmarks such as LostAndFound \cite{pinggera2016lost} (1 landform, 9 anomaly types), Fishyscapes \cite{blum2021fishyscapes} (1 landform, 7 anomalies), and SMIYC-RoadAnomaly21 \cite{chan2021segmentmeifyoucan} (4 landforms, 26 anomalies), ClimaOoD provides substantially broader coverage with 6 landforms, 93 anomaly categories, and 6 distinct weather conditions. This wide diversity ensures more robust training for anomaly segmentation models under varied open-world conditions.

We conducted comprehensive experiments on ClimaOoD by training four state-of-the-art anomaly segmentation methods on our constructed training dataset and testing them on our filtered test set.
Quantitative results demonstrate that incorporating ClimaOoD into training robustly improves anomaly segmentation performance. Across four state-of-the-art methods, AUROC increases by 0.66\%, and AP improves by 3.25\%. However, when evaluated under adverse conditions, models still show robustness gaps, with the average FPR95—rising from 7.8\% in clear weather to 11.0\%, underscoring the need for more realistic OoD data to improve generalization in open-world anomaly detection.
Our contributions are summarized as follows:  
\begin{itemize}

 \item We build ClimaOoD, a comprehensive benchmark encompassing six representative driving scenarios across diverse weather conditions, including clear, foggy, rainy, and snowy environments. It contains over 10,000 high-fidelity image–mask pairs.

 \item We propose ClimaDrive, a semantics-guided image-to-image framework that generates realistic anomalous driving scenes.

\item Extensive experiments show that training with ClimaOoD improves performance across four state-of-the-art methods, boosting average AP by 3.25\% and enhancing model precision and recall across diverse scenes.

\end{itemize}
\section{Related Work \label{sec:relate work}}

\subsection{Anomaly Segmentation Benchmarks in Autonomous Driving}

Anomaly segmentation relies heavily on high-quality and diverse datasets\cite{zhang2023improving,zhao2024segment,lis2019detecting,shoeb2025outofdistributionsegmentationautonomousdriving}. Yet real-world anomaly collection remains difficult—anomalous events are rare, unpredictable, and highly variable across environments\cite{Liu2025OoDDINOAM,Zheng2025SegmentingOA}. This scarcity limits model generalization to diverse out-of-distribution (OoD) scenarios\cite{kim2023sanflow,sun2025anomalyany,zhou2023pad}.

Early benchmarks such as LostAndFound\cite{pinggera2016lost} and Fishyscapes\cite{blum2021fishyscapes} focus mainly on clear-weather urban scenes, offering limited anomaly and scene diversity. RoadAnomaly\cite{lis2019detecting} and SMIYC\cite{chan2021segmentmeifyoucan} expand scene types and weather conditions but remain small in scale. ComsAmy\cite{xia2025opensetanomalysegmentationcomplex} increases scenario and weather coverage yet contains only 468 images, restricting robust evaluation. Overall, existing datasets are too limited or too small to reflect the complexity of real-world driving environments.

These limitations underscore the need for a large-scale, diverse benchmark that spans varied scenarios, weather conditions, and anomaly types—providing a more realistic foundation for developing and evaluating anomaly segmentation models under open-world conditions.

\subsection{Synthetic Data Generation Methods}
In anomaly segmentation, generating synthetic data has become a crucial strategy to overcome the limitations of real-world datasets\cite{Tian2022pixelwise,Lis2020detecting,Vojir2021road,shoeb2025outofdistributionsegmentationautonomousdriving}. Two primary methods have been widely adopted: the \textit{copy-paste} approach and \textit{text-to-image diffusion models}.

The \textit{copy-paste} method inserts objects from external datasets into driving scenes\cite{Tian2022pixelwise,liu2023residual,nayal2023rba,Liu2025OoDDINOAM}, as used in Fishyscapes-Static\cite{blum2021fishyscapes} and StreetHazards\cite{steinhardt2019benchmark}. While it enables easy anomaly diversification, it often breaks contextual consistency—pasted objects frequently mismatch scene geometry, scale, or lighting.
\textit{Text-to-image diffusion} synthesizes anomalies directly via prompts\cite{nie2024compositional,ohanyan2024zero,liang2024rich,loiseau2024reliability}. Loiseau et al.\cite{loiseau2024reliability}, for instance, generate anomalies from randomly sampled boxes. Despite improved visual diversity, these methods lack physical grounding, leading to spatially or geometrically misaligned anomalies.

Overall, existing synthetic approaches enhance diversity but still struggle to ensure semantically coherence and physical realism, motivating the need for a more unified and physically consistent generation framework.

\begin{table*}[t]
\centering
\caption{Comparison of anomaly segmentation benchmarks. 
ClimaOoD provides broader scene coverage, richer weather diversity, and more anomaly types than existing datasets.}
\vspace{-2mm}
\resizebox{\linewidth}{!}{
\begin{tabular}{lccccc}
\toprule
\textbf{Dataset} & \textbf{Anomaly pixels} & \textbf{Size} & \textbf{Landforms} & \textbf{Anomalies} & \textbf{Weather} \\
\midrule
LostAndFound~\cite{pinggera2016lost} & 0.12\% & 1203 & 1 & 9  & clear \\
Fishyscapes LAF~\cite{blum2021fishyscapes} & 0.23\% & 373  & 1 & 7  & clear \\
SMIYC-RoadObstacle21~\cite{chan2021segmentmeifyoucan} & 0.12\% & 327  & 2 & 31 & clear, frosty, night \\
RoadAnomaly~\cite{lis2019detecting} & 9.85\% & 60   & 3 & 21 & clear \\
SMIYC-RoadAnomaly21~\cite{chan2021segmentmeifyoucan} & 13.83\% & 100  & 4 & 26 & clear \\
\midrule
\textbf{ClimaOoD (ours)} & 2.37\% & 1200 & 6 & 93 & clear, rain, fog, snow, cloudy, night \\
\bottomrule
\end{tabular}}
\label{tab:benchmark_comparison}

\end{table*}

\begin{figure*}[t]
    \centering
    \includegraphics[width=\textwidth]{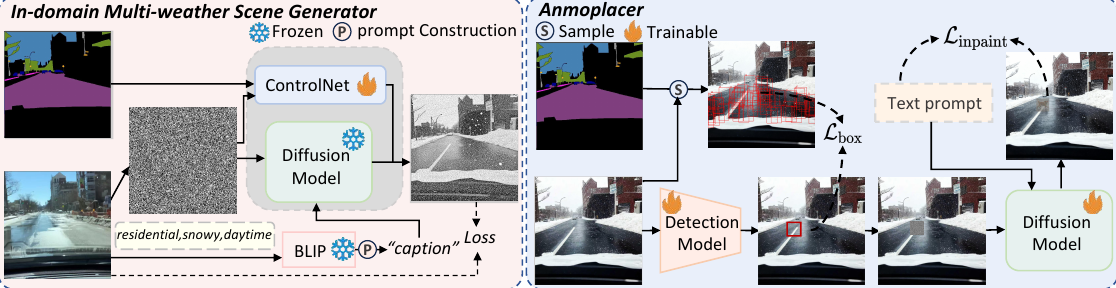}
    \caption{
       ClimaDrive Training Pipeline.
(a) Multi-Scene Weather Generator: A structure-guided diffusion model creates semantically consistent driving scenes under diverse weather conditions using BDD100K inputs and scene-aware prompts.
(b) AnomPlacer: A trainable module that predicts anomaly locations and synthesizes OoD objects via text-conditioned diffusion inpainting (e.g., “dog”).
    }
  \vspace{-5mm}
    \label{fig:training_process}
\end{figure*}

\section{The ClimaOoD Dataset}
\label{sec:climaood_dataset}
This section introduces the proposed \textbf{ClimaDrive} framework and the resulting \textbf{ClimaOoD} dataset.
We first detail the data generation methodology, including the synthesis of realistic multi-weather scenes and physically consistent anomalies.
Then, we present the composition and statistics of ClimaOoD, highlighting its scale, scenario diversity, and comprehensive weather coverage.

\subsection{Generation Framework: ClimaDrive}
ClimaDrive introduces a unified, semantics-guided image-to-image framework that fundamentally redefines out-of-distribution( OoD) data synthesis. Instead of heuristically injecting anomalies or relying on unconstrained inpainting, ClimaDrive enforces perspective-aware spatial validity and semantics-consistent scene rendering as first-class constraints in the generation process. This design upgrades anomaly synthesis from ad-hoc manipulation to a physically grounded generative pipeline that achieves controllable weather variation, reliable spatial placement, and scene-faithful integration—capabilities that prior approaches fundamentally cannot realize. The overall training process is illustrated in Figure~\ref{fig:training_process}.

\noindent\textbf{Multi-scene weather generator:} To enrich environmental diversity while preserving spatial structure, we generate multi-weather driving scenes conditioned on both semantics and text.  
Given a semantic map $S_{\text{sem}} \in \mathbb{R}^{H \times W}$, an image $\tilde{I}$ is synthesized as: $ \tilde{I} \sim p_\theta(I \mid S_{\text{sem}}, p)$
where $p$ is a textual prompt describing scene attributes.

Scene attributes $\alpha = \{\alpha^{\text{weather}}, \alpha^{\text{scene}}, \alpha^{\text{time}}\}$ are derived from BDD100K\cite{yu2020bdd100k} metadata and fused with a caption $c$ from BLIP~\cite{li2022blip} to form $p = f(c, \alpha)$. 
Stable Diffusion~\cite{rombach2022highresolution} with ControlNet~\cite{lv2023controlnet} is employed, where ControlNet is fine-tuned using $S_{\text{sem}}$ as structural guidance and $p$ as the linguistic condition. 
This preserves geometric alignment while achieving realistic variation across weather and lighting conditions.

\noindent\textbf{AnomPlacer:} Anomalous objects in driving scenes must respect both perspective and drivable constraints to ensure realistic placement. To address this, we first extract drivable regions \( R \) from the given semantic layout \( S_{\text{sem}} \), and then sample \( N \) pseudo-bounding boxes \( B = \{b_i = (x_i, y_i, w_i, h_i)\} \), with \( N = 64 \). A perspective prior is applied to adjust the scales of the boxes based on the image height \( H \), ensuring that object size varies according to depth (i.e., the vertical position of the bounding box), as given by the equation: $h_i = \frac{H}{y_i}, \quad w_i \propto h_i.$
This adjustment ensures that anomalies are placed according to the perspective, maintaining physical realism.

Next, a detection backbone \( F_\theta \) predicts the adjusted bounding boxes \( \hat{B} \), which are supervised by Hungarian-matched pseudo boxes \( B \). The localization loss function \( \mathcal{L}_{\text{box}} \) optimizes both the positional alignment and the intersection-over-union (IoU) between predicted and ground truth boxes:

\[
\mathcal{L}_{\text{box}} = \sum_{j=1}^N \Big[\|(x_j,y_j)-(\hat{x}_{\pi(j)},\hat{y}_{\pi(j)})\|_1 + \text{IoU}(b_j,\hat{b}_{\pi(j)})\Big].
\]

To generate anomalous objects, we use a diffusion model for inpainting. Each predicted bounding box \( \hat{b}_j \) is filled with an anomalous object \( \tilde{O}_j \) based on the scene's global context \( S_{\text{scene}} \) (e.g., "Tunnel, Rainy, Daytime") and the object concept \( t_j \) (e.g., “sofa”, “dog”). The diffusion model generates an object \( \tilde{O}_j \) conditioned on the box and the scene: $\tilde{O}_j \sim p_\theta(O \mid \hat{b}_j, S_{\text{scene}}, t_j).$
This ensures that the generated objects are both semantically coherent (appropriate for the scene) and physically consistent (matching perspective and scale).

The overall objective function combines the localization loss \( \mathcal{L}_{\text{box}} \) and the inpainting loss \( \mathcal{L}_{\text{inpaint}} \), which is optimized in two stages. The first stage pretrains the localization module, and the second stage refines the results through joint optimization with the inpainting model: 

\[
\mathcal{L}_{\text{total}} = \mathcal{L}_{\text{box}} + \mathcal{L}_{\text{inpaint}}
\]

Additionally, grounding-based anomaly masks \( M_{\text{ano}} \) are generated using the Grounding-SAM method\cite{zhu2024grounded}, and these masks are further refined using a lightweight noise-denoise process to improve boundary smoothness and coherence, ensuring that the generated anomalies align well with the scene structure.

\subsection{Dataset Construction and Selection}

To enable robust training and evaluation, the ClimaOoD benchmark is constructed through a two-stage process and organized along scene–weather dimensions with consistent anomaly annotations.

First, we generate a large-scale training dataset of 10,230 synthetic images (with corresponding masks), covering 6 weather conditions (clear, rain, fog, snow, cloudy, night) and 6 driving scene types (city street, highway, tunnel, gas station, residential, parking lot). This dataset incorporates 93 anomaly categories—featuring diverse object appearances, scales, and types (e.g., animals, vehicles, obstacles)—laying a solid foundation for model training.

From this extensive training set, we curated a representative 1,200-image test set through manual screening that serves as the core of the ClimaOoD benchmark for evaluation. The selection process prioritized samples with clear semantic annotations, balanced distribution across scene-weather combinations, and realistic representation of anomalous scenarios to ensure benchmark reliability. Table~\ref{tab:benchmark_comparison} highlights that test set markedly broadens scene types, weather conditions, and anomaly diversity, offering a more comprehensive benchmark for evaluating open-world robustness.


Each sample includes a binary anomaly mask for pixel-level assessment. Anomalies occupy 2.37\% of image pixels on average, ranging from small objects to large obstacles, and are placed in perspective-aware drivable regions to ensure realistic geometric integration.

\subsection{Visualization and Statistical Analysis}

\begin{figure}[t]
    \centering
    \includegraphics[width=1\linewidth]{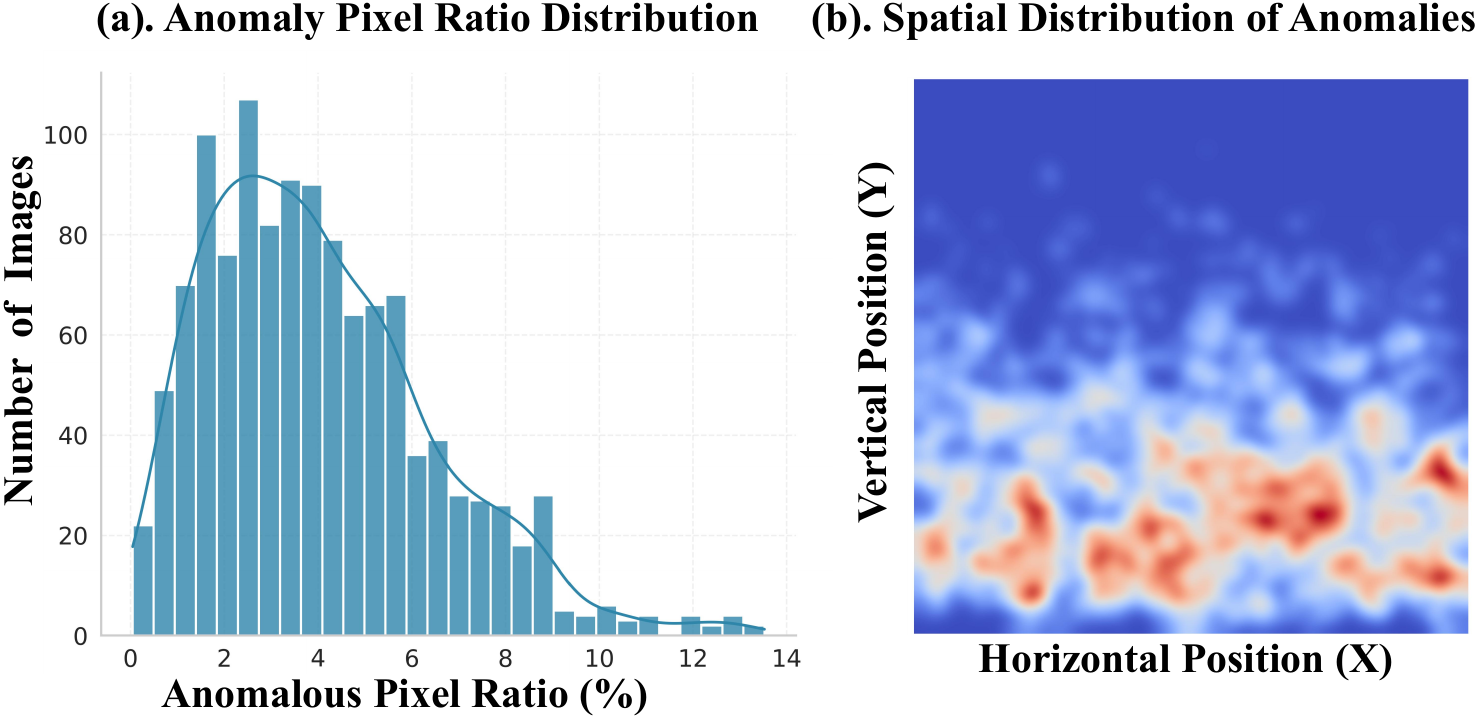}
    \caption{Anomaly pixel statistics in \textbf{ClimaOoD}: (a) Distribution of anomaly pixel fraction per image, indicating scale diversity; (b) Spatial distribution heatmap of anomaly pixels, showing that anomalies frequently appear near drivable regions.}
    \vspace{-5mm}
    \label{fig:anomaly_plots}
\end{figure}

The diversity and scale of ClimaOoD are further demonstrated through visual and statistical analysis. Figure~\ref{fig:anomaly_plots}.(a) shows the distribution of anomaly pixel fractions, highlighting the variability in the scale of anomalies, from small objects to large distractors. Figure~\ref{fig:anomaly_plots}.(b) presents a spatial heatmap that illustrates the concentration of anomalies in drivable areas, reflecting real-world risk zones commonly encountered in autonomous driving.

\begin{figure*}[t]
    \centering
    \vspace{-3mm}
    \includegraphics[width=1\textwidth]{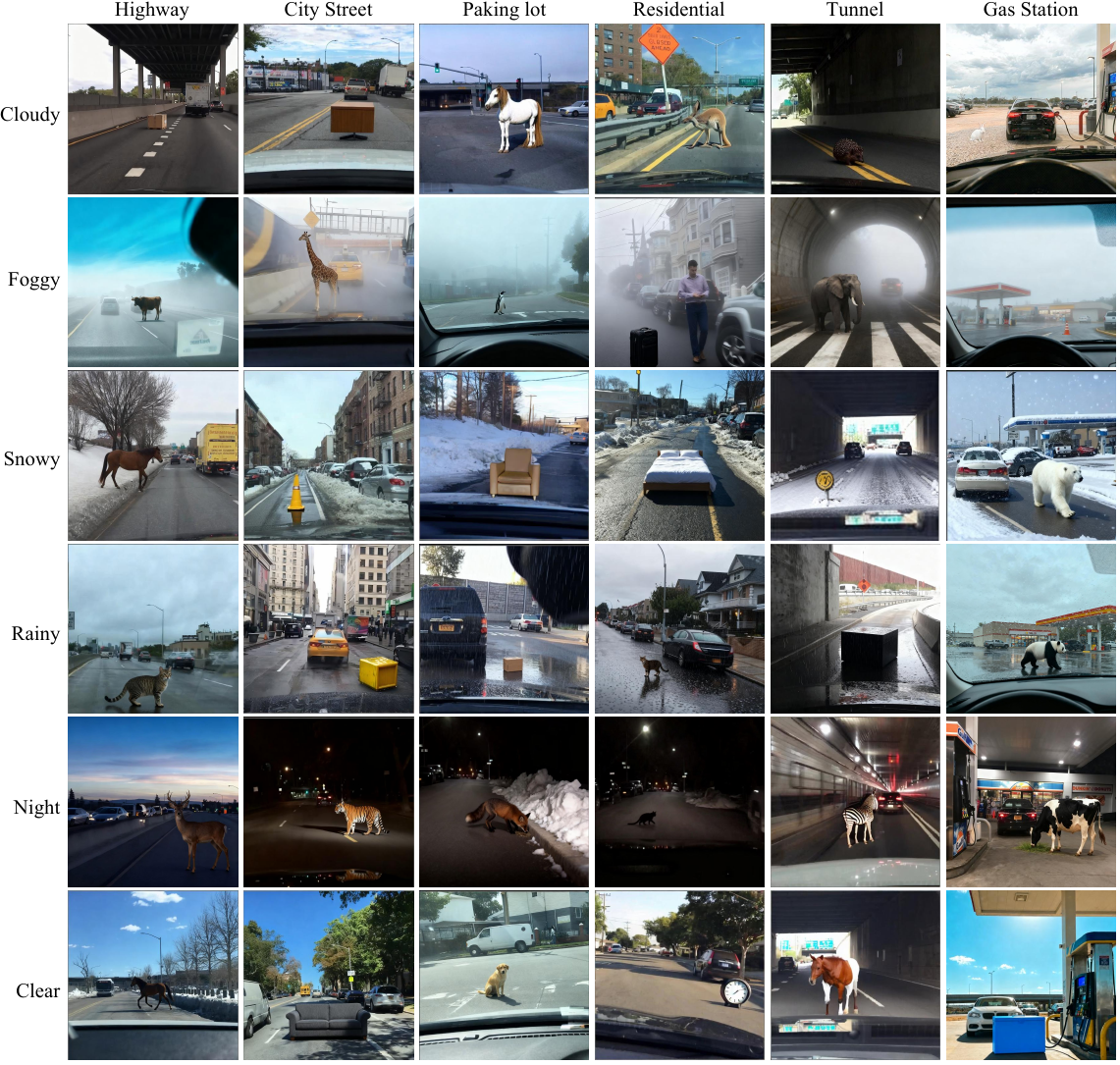}
    \vspace{-5mm}
    \caption{Visual examples from our \textbf{ClimaOoD} dataset showing synthetic anomaly scenes under diverse weather conditions. Each scene exhibits realistic illumination, texture variations, and weather-specific visual cues. Anomalous objects are seamlessly integrated into the environment, highlighting the controllable diversity and fidelity of our generation pipeline.}
    \label{fig:multi_scenario_weather}
    \vspace{-5mm}
\end{figure*}

In addition, Figure~\ref{fig:multi_scenario_weather} presents a $6 \times 6$ visualization grid, showcasing different scene and weather combinations. Each image demonstrates the fidelity and realism of ClimaOoD, with anomalies such as animals, boxes, and other objects seamlessly integrated into various weather conditions like foggy tunnels or rainy highways. This grid illustrates the effectiveness of our generation pipeline in capturing the complexity and diversity of open-world driving scenarios.

\section{Experiment \label{sec:experiment}}
\begin{table*}[t]
\centering
\caption{Comparison of OoD detection performance with and without \textbf{ClimaOoD} as training data. 
Results are reported in AUROC ($\uparrow$), AP ($\uparrow$), and FPR95 ($\downarrow$) in \%. 
Bold indicates better performance between the two training strategies.}
\resizebox{0.85\textwidth}{!}{
\begin{tabular}{l|l|ccc|ccc}
\toprule
\multirow{2}{*}{Method} & \multirow{2}{*}{Dataset} & \multicolumn{3}{c|}{Original Output} & \multicolumn{3}{c}{Trained with ClimaOoD (Ours)} \\
&  & AUROC$\uparrow$ & AP$\uparrow$ & FPR95$\downarrow$ & AUROC$\uparrow$ & AP$\uparrow$ & FPR95$\downarrow$ \\ 
\midrule
\multirow{4}{*}{RbA\cite{nayal2023rba}} 
& FishyScape LAF        & 97.19 & 78.27 & 3.97 & \textbf{97.85} & \textbf{81.32} & \textbf{3.52} \\
& RoadAnomaly            & 98.16 & 88.62 & 6.07 & \textbf{98.53} & \textbf{90.47} & \textbf{5.21} \\
& RoadAnomaly21        & 97.61 & 89.39 & 8.42 & \textbf{98.14} & \textbf{91.26} & \textbf{7.35} \\
& RoadObstacle21        & 99.81 & \textbf{98.40} & 0.04 & \textbf{99.86} & 97.95 & \textbf{0.03} \\ 
\midrule
\multirow{4}{*}{RPL\cite{liu2023residual}} 
& FishyScape LAF        & 99.38 & 70.62 & \textbf{2.52} & \textbf{99.51} & \textbf{75.89} & 2.78 \\
& RoadAnomaly            & 95.72 & 71.60 & 17.75 & \textbf{97.26} & \textbf{78.93} & \textbf{14.32} \\
&  RoadAnomaly21         & 98.05 & 88.53 & 7.22 & \textbf{98.71} & \textbf{91.08} & \textbf{5.98} \\
& RoadObstacle21        & 99.96 & 96.71 & \textbf{0.09} & \textbf{99.97} & \textbf{97.54} & 0.12 \\ 
\midrule
\multirow{4}{*}{Mask2Anomaly\cite{rai2024mask2anomaly}}
& FishyScape LAF        & 95.41 & 69.46 & 9.31 & \textbf{96.28} & \textbf{74.15} & \textbf{7.86} \\
& RoadAnomaly            & 96.54 & 80.05 & 13.94 & \textbf{97.32} & \textbf{83.67} & \textbf{11.59} \\
& RoadAnomaly21        & \textbf{98.93} & \textbf{95.49} & 2.39 & 98.87 & 94.96 & \textbf{2.18} \\
& RoadObstacle21        & 99.91 & 92.83 & 0.16 & \textbf{99.93} & \textbf{95.26} & \textbf{0.13} \\ 
\midrule
\multirow{4}{*}{UNO\cite{delić2024outlierdetectionensemblinguncertainty}}
& FishyScape LAF        & 96.88 & 74.49 & 6.86 & \textbf{97.63} & \textbf{80.12} & \textbf{5.79} \\
& RoadAnomaly            & \textbf{97.71} & 87.21 & 6.88 & 97.54 & \textbf{89.76} & \textbf{5.42} \\
& RoadAnomaly21        & \textbf{97.87} & 90.28 & 6.02 & 97.73 & \textbf{92.04} & \textbf{4.85} \\
& RoadObstacle21        & \textbf{99.58} & 80.09 & 1.59 & 99.46 & \textbf{84.37} & \textbf{1.38} \\ 
\bottomrule
\end{tabular}
}
\label{tab:climaood_results}
\end{table*}

Our experiments employ ClimaOoD for training and test set for benchmarking four state-of-the-art anomaly segmentation models.

\noindent\textbf{Methods.}
These four methods represent two main paradigms in out-of-distribution(OoD) anomaly segmentation. Outlier-exposure methods (RPL\cite{liu2023residual}, Mask2Anomaly\cite{rai2024mask2anomaly}) aim to directly highlight abnormal regions: RPL\cite{liu2023residual} learns residual patterns to increase pixel-level sensitivity, while Mask2Anomaly\cite{rai2024mask2anomaly} formulates anomaly detection as mask classification with masked attention.
Uncertainty-based methods (RbA\cite{nayal2023rba}, UNO\cite{delić2024outlierdetectionensemblinguncertainty}) calibrate anomaly scores by modeling prediction confidence: RbA\cite{nayal2023rba} suppresses overconfident in-distribution logits, and UNO\cite{delić2024outlierdetectionensemblinguncertainty} adopts a $K{+}1$ classifier with an explicit outlier class to refine anomaly scoring.

\noindent\textbf{Evaluation Metrics.} 
We adopt three standard OoD detection metrics: Average Precision (AP) measures overall detection accuracy, Area Under the ROC Curve (AUROC) evaluates the model’s ability to distinguish in- and out-of-distribution pixels, and FPR95 denotes the false positive rate when the true positive rate is 95\%, reflecting detection reliability.

\subsection{Training OoD Segmentation Models}
To assess the impact of training with the ClimaOoD dataset, we compare the performance of four state-of-the-art OoD segmentation models when trained with ClimaOoD versus original datasets. The models are evaluated on four benchmark datasets (Fishyscapes LAF\cite{blum2021fishyscapes}, RoadAnomaly\cite{lis2019detecting}, RoadAnomaly21\cite{chan2021segmentmeifyoucan}, and RoadObstacle21\cite{chan2021segmentmeifyoucan}) to examine the generalization capability under diverse driving scenarios and weather conditions.

\noindent\textbf{Results and Analysis.}
Table~\ref{tab:climaood_results} summarizes the performance improvements when training with ClimaOoD. Across all methods, training with ClimaOoD results in notable gains in AUROC, AP, and FPR95. For example, RbA\cite{nayal2023rba} achieves an increase in AUROC from 97.19 to 97.85 on Fishyscapes LAF, and AP improves from 78.27 to 81.32 on RoadAnomaly. Similarly, Mask2Anomaly\cite{rai2024mask2anomaly} sees improvements in both AUROC (from 95.41 to 96.28) and AP (from 69.46 to 74.15) across the datasets.

In particular, training with ClimaOoD enhances the robustness of these models, with FPR95 decreasing in most cases (e.g., RbA\cite{nayal2023rba} drops from 3.97 to 3.52 on Fishyscapes LAF). These results demonstrate that ClimaOoD provides valuable, diverse training data that leads to better generalization and performance under a wide range of OoD conditions. The improvements highlight the significance of training with realistic, weather-diverse data for improving OoD segmentation in open-world environments.

\subsection{Benchmarking OoD Segmentation Models}
\begin{table*}[t]
\centering
\caption{
OoD detection results across different \textbf{scenarios} and \textbf{weather conditions} on \textbf{ClimaOoD}.
We report AUROC (↑), AP (↑), and FPR95 (↓) for four state-of-the-art methods, with all values in percentage (\%) .
Best results are in \textbf{bold}, second best are \underline{underlined}.
}
\vspace{-2mm}
\resizebox{\linewidth}{!}{
\begin{tabular}{l|l|ccc|ccc|ccc|ccc}
\toprule
\multirow{2}{*}{\textbf{Scenario}} & \multirow{2}{*}{\textbf{Weather}} &
\multicolumn{3}{c|}{\textbf{RPL}~\cite{liu2023residual}} &
\multicolumn{3}{c|}{\textbf{RbA}\cite{nayal2023rba}} &
\multicolumn{3}{c|}{\textbf{Mask2Anomaly}\cite{rai2024mask2anomaly}} &
\multicolumn{3}{c}{\textbf{UNO}\cite{delić2024outlierdetectionensemblinguncertainty}} \\
 & & AUROC & AP & FPR95 & AUROC & AP & FPR95 & AUROC & AP & FPR95 & AUROC & AP & FPR95 \\
\midrule
\multirow{2}{*}{CityStreet} 
  & Clear    & \textbf{98.10} & \underline{74.00} & 6.80 & 94.50 & 50.00 & 23.00 & 95.20 & 60.00 & 21.00 & \underline{98.00} & \textbf{74.80} & \textbf{6.00} \\
  & Adverse  & \textbf{97.80} & \underline{72.00} & 7.40 & 94.00 & 48.00 & 25.00 & 94.80 & 58.50 & 22.00 & \underline{97.60} & \textbf{73.00} & \textbf{6.80} \\
\midrule
\multirow{2}{*}{Gas station} 
  & Clear    & \textbf{98.00} & \underline{80.50} & 11.00 & 94.20 & 52.00 & 21.00 & 95.10 & 61.00 & 20.50 & \underline{98.40} & \textbf{81.50} & \textbf{10.20} \\
  & Adverse  & \textbf{97.60} & \underline{79.00} & 11.80 & 93.80 & 50.00 & 22.50 & 94.70 & 59.00 & 21.50 & \underline{98.10} & \textbf{80.00} & \textbf{11.00} \\
\midrule
\multirow{2}{*}{Highway} 
  & Clear    & \textbf{98.90} & \underline{82.80} & \textbf{4.50} & 95.00 & 53.00 & 20.00 & 95.50 & 62.00 & 19.50 & \underline{98.70} & \textbf{83.50} & \underline{4.80} \\
  & Adverse  & \textbf{98.50} & \underline{81.20} & \textbf{5.00} & 94.20 & 51.00 & 22.00 & 95.20 & 60.50 & 20.50 & \underline{98.30} & \textbf{82.00} & \underline{5.40} \\
\midrule
\multirow{2}{*}{ParkingLot} 
  & Clear    & \textbf{99.20} & \underline{87.20} & 4.20 & 94.80 & 51.50 & 21.50 & 95.60 & 62.00 & 20.00 & \underline{99.10} & \textbf{88.00} & \textbf{3.80} \\
  & Adverse  & \textbf{98.90} & \underline{86.00} & 4.80 & 94.00 & 50.50 & 23.00 & 95.10 & 61.00 & 21.00 & \underline{98.70} & \textbf{87.00} & \textbf{4.20} \\
\midrule
\multirow{2}{*}{Residential} 
  & Clear    & \textbf{99.20} & \underline{79.50} & \textbf{2.40} & 94.60 & 52.50 & 20.80 & 95.30 & 62.00 & 20.20 & \underline{99.00} & \textbf{80.20} & \underline{2.70} \\
  & Adverse  & \textbf{98.90} & \underline{78.50} & \textbf{2.80} & 94.20 & 51.00 & 22.00 & 95.00 & 61.00 & 21.00 & \underline{98.70} & \textbf{79.50} & \underline{3.10} \\
\midrule
\multirow{2}{*}{Tunnel} 
  & Clear    & \textbf{97.40} & \underline{69.50} & \textbf{9.80} & 91.80 & 40.00 & 35.00 & 93.00 & 52.00 & 28.00 & \underline{97.20} & \textbf{70.50} & \underline{10.40} \\
  & Adverse  & \textbf{97.00} & \underline{68.00} & \textbf{10.50} & 91.00 & 38.00 & 37.00 & 92.50 & 51.00 & 29.00 & \underline{96.90} & \textbf{69.00} & \underline{11.20} \\
\midrule
\textbf{Average} 
  & All      & 98.30 & 77.30 & 8.00 & 94.00 & 51.00 & 22.00 & 94.90 & 59.00 & 24.10 & \textbf{98.60} & \textbf{80.50} & \textbf{7.80} \\
\bottomrule
\end{tabular}}
\vspace{-3mm}
\label{tab:sota_climaood_weather}
\end{table*}
To evaluate the robustness of OoD segmentation methods in diverse driving environments, we benchmark four representative models on the ClimaOoD's test set, covering various scenarios (e.g., urban streets, highways) and weather conditions (e.g., clear, adverse). All models are tested in their official form, ensuring the focus is on generalization across different environments.

\noindent\textbf{Results and Analysis.}
Table~\ref{tab:sota_climaood_weather} shows that all methods deteriorate under adverse weather compared with clear conditions. For example, RPL’s AUROC drops from 98.10 to 97.80 in \textit{CityStreet}, and the largest declines appear in tunnel and nighttime scenes, where AP decreases by 3–5 points due to occlusion and low illumination.

Although UNO\cite{delić2024outlierdetectionensemblinguncertainty} and RPL\cite{liu2023residual} achieve the strongest overall results, all methods show clear robustness gaps in complex environments. Unlike prior benchmarks such as Fishyscapes\cite{blum2021fishyscapes}—restricted to clear-weather urban scenes—ClimaOoD exposes failure cases across diverse weather and lighting, providing a more realistic and comprehensive evaluation of OoD segmentation performance.

\subsection{Ablation Studies}
\label{sec:ablation}

To assess the impact of ClimaOoD on OoD detection, we conduct ablations on SMIYC-RoadAnomaly21 and SMIYC-RoadObstacle21 using two representative methods, RbA and RPL.
We compare three training configurations of equal size: (1) Original, using copy-paste and text-to-image data; (2) Clear\&City Street, using clear-weather city scenes from ClimaOoD; and (3) Full ClimaOoD, using all weather conditions and driving scenarios.

\noindent\textbf{Results.}
Table~\ref{tab:ablation_single} shows that training on ClimaOoD (Clear) gives moderate gains, while Full ClimaOoD consistently yields the best performance.
On RoadAnomaly21, RbA improves from 97.61 to 98.14 AUROC and reduces FPR95 from 8.42 to 7.35; RPL improves from 98.05 to 98.71 AUROC with FPR95 dropping from 7.22 to 5.98.
On RoadObstacle21, both methods maintain high AUROC (99.8+); slight AP and FPR95 variations stem from the small object sizes in RoadObstacle21, which differ from those in our synthetic data.

\begin{figure}[!t]
\centering
\includegraphics[width=\linewidth]{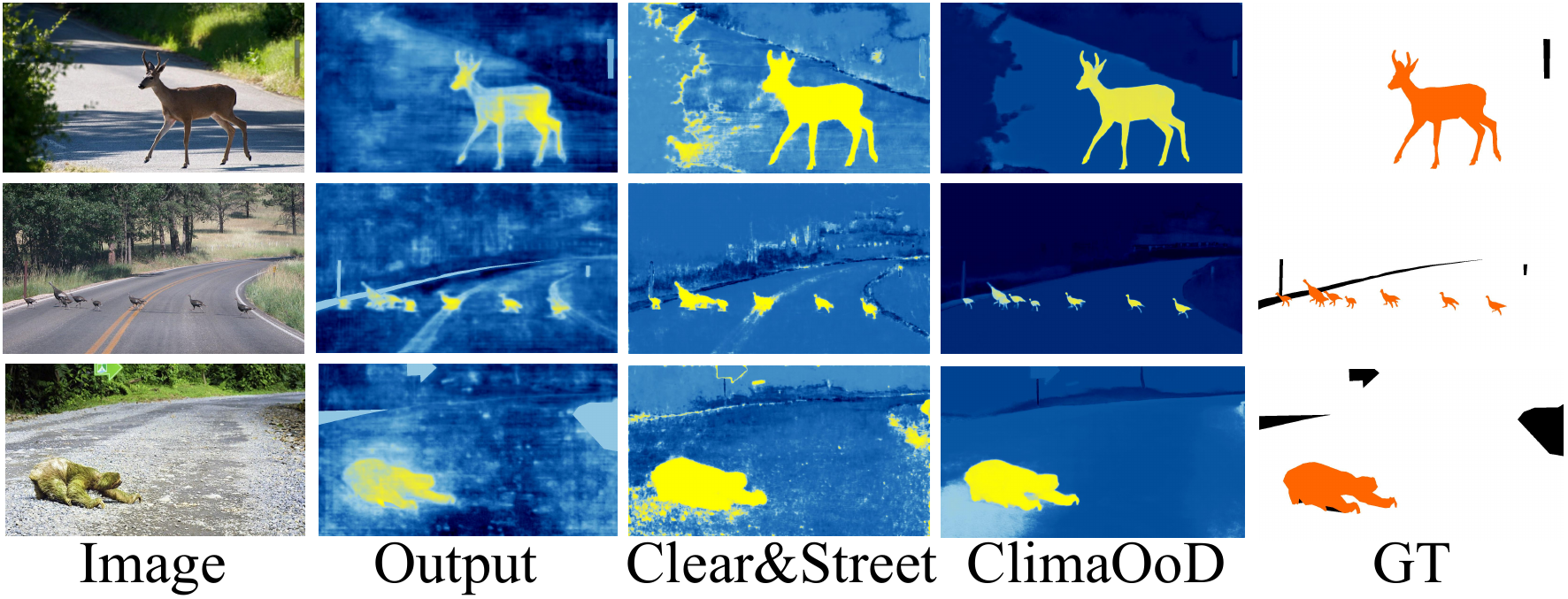}

\caption{Visualization of Ablation Study for OoD Detection: RPL Outputs from Diverse Training Datasets (Left to Right: Image, Original RPL Output, Clear\&Street-trained, ClimaOoD-trained, Ground Truth(GT))}
\label{fig:ablation_vis}
\end{figure}

Qualitative results in Figure~\ref{fig:ablation_vis} further confirm that models trained on ClimaOoD better localize diverse anomalies, producing cleaner and more coherent anomaly maps than those trained on the baseline or clear-only data.

Overall, these results show that ClimaOoD substantially improves OoD robustness by providing richer environmental diversity and more realistic anomaly appearance.

\begin{table}[t]
\centering
\caption{Ablation study on training data for OoD detection. Metrics: AUROC ($\uparrow$), AP ($\uparrow$), FPR95 ($\downarrow$).}
\setlength{\tabcolsep}{1.4mm} 
\renewcommand{\arraystretch}{1} 
\footnotesize  
\begin{tabular}{llccc}
\toprule
Method & Training Set & AUROC$\uparrow$ & AP$\uparrow$ & FPR95$\downarrow$ \\
\midrule
\multicolumn{5}{l}{\textit{RoadAnomaly21}} \\
\midrule
\multirow{3}{*}{RbA\cite{nayal2023rba}} 
& Original & 97.61 & 89.39 & 8.42 \\
& Clear\&City Street  & 97.83 & 90.15 & 7.96 \\ 
& Full ClimaOoD & \textbf{98.14} & \textbf{91.26} & \textbf{7.35} \\
\midrule
\multirow{3}{*}{RPL\cite{liu2023residual}} 
& Original & 98.05 & 88.53 & 7.22 \\
& Clear\&City Street & 98.28 & 89.77 & 6.85 \\ 
& Full ClimaOoD & \textbf{98.71} & \textbf{91.08} & \textbf{5.98} \\
\midrule
\multicolumn{5}{l}{\textit{RoadObstacle21}} \\
\midrule
\multirow{3}{*}{RbA\cite{nayal2023rba}} 
& Original & 99.81 & \textbf{98.40} & 0.04 \\
& Clear\&City Street & 99.84 & 98.27 & 0.06\\ 
& Full ClimaOoD & \textbf{99.86} & 97.95 & \textbf{0.03} \\
\midrule
\multirow{3}{*}{RPL\cite{liu2023residual}} 
& Original & 99.96 & 96.71 & \textbf{0.09} \\
& Clear\&City Street  & 99.80 & 96.93 & 0.15 \\ 
& Full ClimaOoD & \textbf{99.97} & \textbf{97.54} & 0.12 \\
\bottomrule
\end{tabular}
\label{tab:ablation_single}

\end{table}

\subsection{Technical  Analysis}

Table~\ref{tab:ablation_main} presents an analysis of the key components in the ClimaDrive framework. The goal of this analysis is to understand the contribution of each component to the overall performance in generating realistic OoD data. The performance is evaluated using four key metrics: Fréchet Inception Distance (FID), Learned Perceptual Image Patch Similarity (LPIPS), and two Pearson Correlation Coefficients (PCC) based on entropy and max-logit scores, which measure the alignment between \textbf{generated data} and \textbf{real-world} OoD data from the SMIYC-RoadAnomaly21\cite{chan2021segmentmeifyoucan} benchmark, due to the similarity in object scale to our synthetic data. Specifically, entropy and max-logit are used to assess pixel-level uncertainty in model predictions.

The Full Model obtains the best results, with the lowest FID/LPIPS and highest PCC. Removing box supervision weakens physical alignment, and sparse sampling ($N=32$) further degrades placement accuracy. Dense sampling ($N=128$) improves the scores but still underperforms the full model. Overall, each component is necessary for generating physically consistent anomalies and achieving strong alignment with real OoD data.

\begin{table}[t]
\centering
\caption{Ablation on the main components of ClimaDrive. Lower FID/LPIPS and higher PCC indicate better alignment with real OoD data.}
\label{tab:ablation_main}
\vspace{-3mm}
\resizebox{\columnwidth}{!}{
\begin{tabular}{lcccc}
\toprule
Setting & FID ↓ & LPIPS ↓ & \multicolumn{2}{c}{PCC ↑} \\
\cmidrule(lr){4-5}
& & & entropy & max-logit \\
\midrule
Full Model & \textbf{175.2} & \textbf{0.276} & \textbf{0.86} & \textbf{0.87} \\
w/o Box Supervision & 191.3 & 0.293 & 0.78 & 0.80 \\
w/o Joint Optimization & 188.5 & 0.287 & 0.79 & 0.81 \\
w/o Perspective Prior & 190.2 & 0.291 & 0.77 & 0.79 \\
Sparse Sampling ($N{=}32$) & 183.9 & 0.282 & 0.80 & 0.82 \\
Dense Sampling ($N{=}128$) & 178.8 & 0.279 & 0.81 & 0.83 \\
\bottomrule
\end{tabular}}
\vspace{-5mm}
\end{table}

Additional, we compare SD1.5\cite{rombach2022high}, SD2\cite{stabilityai_sd2_inpainting}, and SDXL\cite{SDXL:2023} as inpainting backbones under the same training loss. 
Table~\ref{tab:efficiency_diffusion} shows SDXL gives the best quality (FID=164.8, LPIPS=0.263) but is computationally heavy. 
SD2 achieves nearly the same realism with 40\% fewer GFLOPs and is adopted as default.

\begin{table}[t]
\centering
\caption{Comparison of diffusion backbones within ClimaDrive. SD2 balances quality and efficiency.}
\vspace{-3mm}
\label{tab:efficiency_diffusion}
\resizebox{1\columnwidth}{!}{
\begin{tabular}{lcccccc}
\toprule
Backbone & Params (B) & GFLOPs & FID ↓ & LPIPS ↓ & \multicolumn{2}{c}{PCC (↑)} \\
\cmidrule(lr){6-7}
 & & & & & Entropy & Max-logit \\
\midrule
SD1.5 & 0.89 & 185 & 186.5 & 0.291 & 0.72 & 0.74 \\
SD2 (Default) & 1.25 & 240 & 175.2 & 0.276 & 0.86 & 0.87 \\
SDXL & 2.60 & 410 & \textbf{164.8} & \textbf{0.263} & \textbf{0.88} & \textbf{0.89} \\
\bottomrule
\end{tabular}}
\vspace{-3mm}
\end{table}

To verify universality, we replace the detection module with DETR\cite{carion2020end}-ResNet50 and DINO\cite{zhang2022dino}-Swin-B, trained jointly with SD2. 
As shown in Table~\ref{tab:ablation_backbone}, DETR yields FID=178.4 and PCC$_\text{entropy}$=0.84, while DINO slightly improves to 175.2 and 0.86. 
The difference across all metrics is under 2\%, confirming stable performance across architectures.

\begin{table}[t]
\centering
\caption{Different detection backbones in AnomPlacer. Results confirm geometric supervision.}
\label{tab:ablation_backbone}
\resizebox{1\columnwidth}{!}{
\begin{tabular}{lcccccc}
\toprule
Backbone & Params (B) & GFLOPs & FID ↓ & LPIPS ↓ & \multicolumn{2}{c}{PCC (↑)} \\
\cmidrule(lr){6-7}
 & & & & & Entropy & Max-logit \\
\midrule
DETR (ResNet-50) & 1.20 & 250 & 178.4 & 0.279 & 0.84 & 0.85 \\
DINO (Swin-B) & 1.25 & 240 & \textbf{175.2} & \textbf{0.276} & \textbf{0.86} & \textbf{0.87} \\
\bottomrule
\end{tabular}}
\end{table}

\begin{figure}[t] \centering \includegraphics[width=1\linewidth]{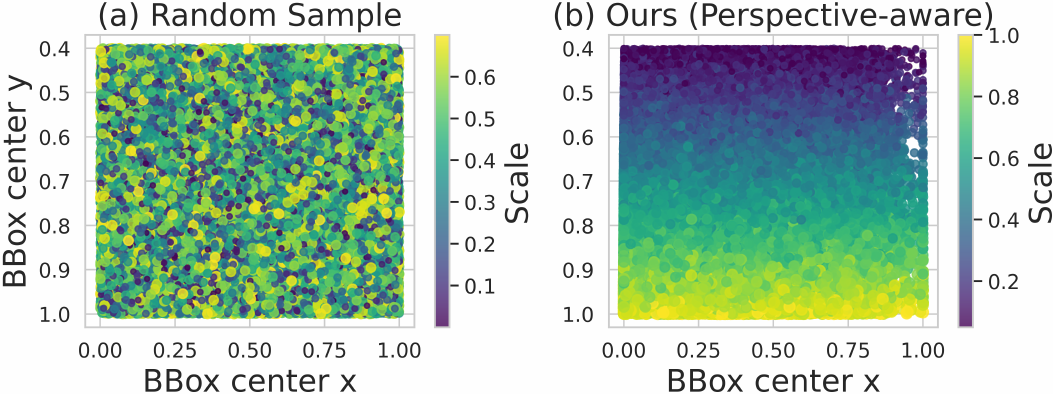} \caption{ Comparison of box distributions shows that random sampling leads to scattered placements, while our perspective-aware sampling creates a natural depth–size pattern, enhancing anomaly location prediction.} \label{fig:box_distribution} 
\vspace{-5mm}
\end{figure}

As shown in Figure~\ref{fig:box_distribution}, random sampling produces physically implausible placements, with box locations and scales showing no spatial correlation. In contrast, our perspective-aware sampling enforces a natural depth–size relationship, where boxes closer to the ego-vehicle appear larger. This simple geometric prior provides an effective pseudo-supervision signal, enabling more realistic and consistent anomaly placement without extra annotations.


\section{Conclusion \label{sec:conclusion}}

In this work, we introduce ClimaOoD, a comprehensive benchmark with over 10,000 high-quality image–mask pairs across six driving scenarios and various weather conditions, enabling robust anomaly segmentation evaluation. ClimaDrive, our proposed framework, integrates AnomPlacer for consistent object placement and a weather generator for diverse environmental synthesis.
Experiments show that training with ClimaOoD improves model performance, highlighting its effectiveness in enhancing precision and recall in anomaly segmentation.

\section{Acknowledgments}
This research was partially funded by the National Natural Science Foundation of China through grants 62402482 and 61902015. Additional support was provided by the CCF-Huawei Populus Grove Fund under grant CCF-HuaweiSE202407, as well as the Fundamental Research Funds for the Central Universities, specifically those of Southwest Minzu University, under award number ZYN2025010.

{
    \small
    \bibliographystyle{ieeenat_fullname}
    \bibliography{main}
}



\end{document}